\documentclass[preprint,12pt]{elsarticle}

\usepackage{graphicx}
\usepackage{amssymb}

\usepackage{amsmath,graphicx, caption, subcaption, amssymb, algorithm, mathrsfs, lipsum, url}

\journal{Optik}

\begin{document}

\begin{frontmatter}


\title{Application of Independent Component Analysis Techniques in Speckle Noise Reduction of Retinal OCT Images}



\author{Ahmadreza Baghaie$^1$, Roshan M. D'Souza$^2$, Zeyun Yu$^3$}

\address{$^1$Department of Electrical Engineering, $^2$Department of Mechanical Engineering, $^3$Department of Computer Science,\\ University of Wisconsin-Milwaukee, Milwaukee, WI, USA}

\begin{abstract}
Optical Coherence Tomography (OCT) is an emerging technique in the field of biomedical imaging, with applications in ophthalmology, dermatology, coronary imaging etc. OCT images usually suffer from a granular pattern, called speckle noise, which restricts the process of interpretation. Therefore the need for speckle noise reduction techniques is of high importance. To the best of our knowledge, use of Independent Component Analysis (ICA) techniques has never been explored for speckle reduction of OCT images. Here, a comparative study of several ICA techniques (InfoMax, JADE, FastICA and SOBI) is provided for noise reduction of retinal OCT images. Having multiple B-scans of the same location, the eye movements are compensated using a rigid registration technique.  Then, different ICA techniques are applied to the aggregated set of B-scans for extracting the noise-free image.
Signal-to-Noise-Ratio (SNR), Contrast-to-Noise-Ratio (CNR) and Equivalent-Number-of-Looks (ENL), as well as analysis on the computational complexity of the methods, are considered as metrics for comparison. The results show that use of ICA can be beneficial, especially in case of having fewer number of B-scans.
\end{abstract}

\begin{keyword}
Independent Component Analysis \sep Speckle Reduction \sep Optical Coherence Tomography (OCT)


\end{keyword}

\end{frontmatter}



\section{Introduction}
\label{S:1}

OCT is a powerful imaging system for non-invasive acquisition of 3D volumetric images of tissues \cite{huang1991optical}, with applications in ophthalmology, dermatology, coronary imaging etc. Due to its underlying physics, which is common in narrow-band detection systems like Synthetic-Aperture Radar (SAR) and ultrasound, OCT images usually suffer from a granular pattern called \textit{speckle}.  Not only the optical properties of the system, but also the motion of the subject to be imaged, size and temporal coherence of the light source, multiple scattering, phase deviation of the beam and aperture of the detector can affect the speckle \cite{schmitt1999speckle}. Fig. 1 shows a sample retinal OCT image, highly degraded by speckle noise.

Speckle is considered to be multiplicative noise, in contrast to the additive Gaussian noise. Limited dynamic range of displays requires us to compress the OCT signals usually by a logarithmic transform, which converts the multiplicative speckle noise to additive noise \cite{salinas2007comparison}. Two major classes of speckle noise reduction techniques are: 1) methods of noise reduction during the acquisition time and 2) post-processing techniques. In the first class multiple uncorrelated recordings are averaged. This includes spatial \cite{avanaki2013spatial}, angular \cite{schmitt1997array}, polarization \cite{kobayashi1991polarization} and frequency \cite{pircher2003speckle} compounding techniques. As for post-processing, anisotropic diffusion-based techniques \cite{salinas2007comparison} and multi-scale/multi-resolution geometric representation techniques \cite{pizurica2008multiresolution} are of high interest between scholars. Use of compressive sensing and sparse representation have also been explored in the past few years \cite{fang2013fast}. For a more complete review on the different image analysis techniques in OCT image processing, including noise reduction, the reader is referred to \cite{baghaie2014state} and references therein. 

Post-processing averaging/median filtering is also an interesting method for speckle reduction. In such techniques, multiple B-scans of the same location are acquired and then the average/median is taken. The misalignment between the different B-scans is usually compensated with a parametric image registration technique. Theoretically,  having $ N $ B-scans with uncorrelated speckle,  SNR can be improved by a factor of  $ \sqrt{N} $.  The works presented in \cite{jorgensen2007enhancing, alonso2011speckle}  can be mentioned as examples. Recently the use of sparse and low-rank decomposition based batch image alignment was explored by the authors \cite{baghaie2014sparse}.

In this paper the use of Independent Component Analysis (ICA) techniques for speckle noise reduction of retinal OCT images is explored, which to the best of our knowledge has never been investigated before. Having multiple B-scans of the same location in retina, the eye movement is compensated by considering a rigid transformation between consecutive B-scans using ImageJ \cite{schneider2012nih}. Having negligible eye motion within each B-scan, the need for deformable registration techniques  \cite{baghaie2014fast, baghaie2014curvature} can be eliminated. Then, several ICA techniques are used for extracting the noise-free image from multiple noisy B-scans. SNR, CNR and ENL are considered as metrics for comparing the performance of different methods. Investigating the results reveals interesting facts regarding the impact of using ICA for speckle noise reduction. Especially in case of having a few number of images in which ICA techniques provide better performance in comparison to median filtering. But increasing the number of images causes the ICA techniques to perform poorly, since the main assumption in ICA is having non-Gaussian uncorrelated components which can't be completely satisfied here.

\begin{figure}
\centering
\includegraphics[scale=.65]{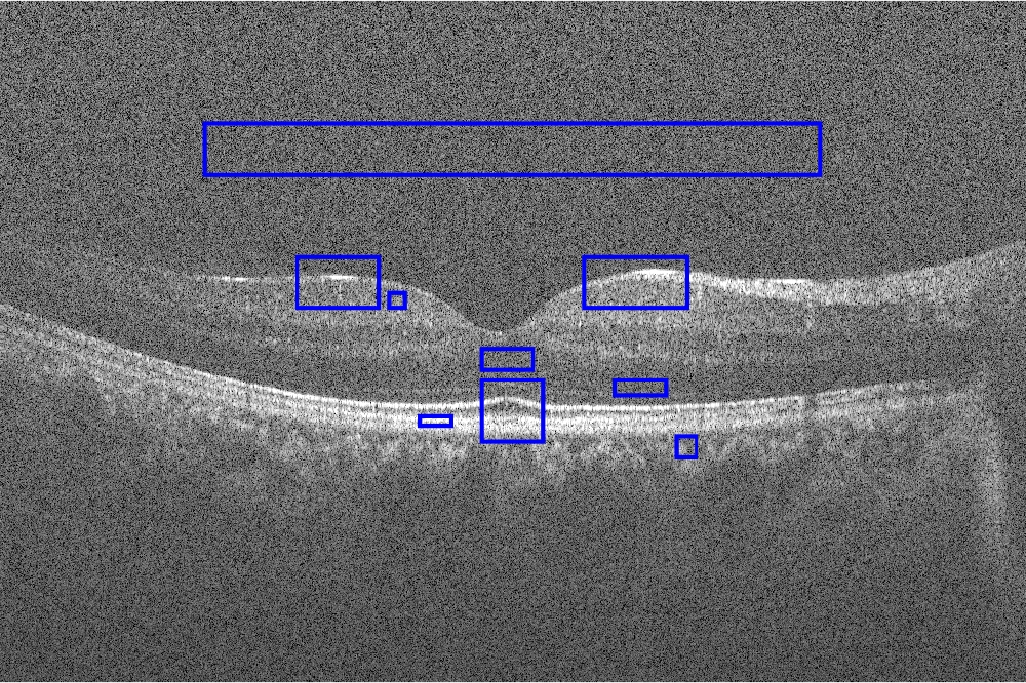}
\caption{Sample retinal OCT image degraded by speckle noise; selected ROIs are shown with blue rectangles.}
\end{figure}

\section{Independent Component Analysis (ICA)}
\label{S:2}

ICA is one of the most widely used techniques for Blind Source Separation (BSS) \cite{naik2011overview}. The problem of BSS consists of having interfering signals from multiple sources recorded and trying to find the individual source signals from these mixed recordings. The very well-known example is a cocktail party, with multiple people talking while there are recorders in different places of the room. This can be mathematically modeled by considering $s_i(t), i=1,...,N$ as the set of sources and $x_i(t), i=1,...,N$ as the observations. Therefore:
\begin{equation}
\mathbf x(t)=\mathbf A \mathbf s(t)
\end{equation}
where $\mathbf A$ is the unknown \textit{mixing matrix}. Recovering the original signals $s_i(t)$ from the recordings $x_i(t)$ is the objective of BSS. Therefore, estimating the unmixing matrix $\mathbf W$ such that $\mathbf W=\mathbf A^{-1}$, the sources can be estimated:
\begin{equation}
\hat{\mathbf s}(t)=\mathbf W \mathbf x(t)
\end{equation}

But how to find $\mathbf W$ when you are "blind" to the nature of mixing matrix and the source signals themselves? This question results in defining a set of constraints on the nature of source signals. As is obvious from the name, ICA, the components should be \textit{statistically independent}. Mathematically speaking, having two random variables $s_1$ and $s_2$, the joint probability density function (pdf) $p_{1, 2}(s_1, s_2)$ for independent variables can be computed as:
\begin{equation}
p_{1, 2}(s_1, s_2)=p_1(s_1)p_2(s_2)
\end{equation}
where $p_1(s_1)$ and $p_2(s_2)$ are the marginal pdf's. 

Another constraint of the signal sources is that they should be non-Gaussian. Based on the central limit theorem, the distribution of a sum of independent signals with arbitrary distributions tends to be a Gaussian distribution, which basically means any Gaussian distribution can be considered as a sum of multiple independent signals. Therefore having Gaussian signals as components prevents us from being able to distinguish them properly. 

ICA has two ambiguities. The variance and the sign of the independent components cannot be estimated. This is usually insignificant in most of the applications. The second ambiguity is regarding the order of the extracted components which cannot be specified. 

As for pre-processing the data before independent component analysis, centering the data (subtracting the mean value) is usually the first step in order to make the ICA procedure easier. The mean will be added after extracting the components. Whitening the data for having uncorrelated components with unit variances is next. 

Here four well-known ICA methods are used for noise reduction of retinal OCT images: InfoMax (RUNICA) \cite{bell1995information}, FastICA \cite{hyvarinen1999fast}, JADE \cite{cardoso1993blind} and SOBI \cite{belouchrani1993second}. In each one of these techniques a contrast/likelihood function is defined and optimized for achieving the optimal separation between sources. The InfoMax method works based on maximizing the output entropy (information flow) of a neural network with non-linear outputs. FastICA on the other hand tries to maximize the non-Gaussianity using an approximation for negentropy utilizing a fixed-point numerical scheme. Joint Approximate Diagonalisation of Eigen-matrices (JADE) method works based on digonalization of the sample data's fourth-order cumulants of the whitened process created from the sample data and a whitening matrix. On the other hand, Second Order Blind Identification (SOBI) assumes that the sample data is gathered from a set of temporally correlated sources and tries to separate them by joint diagonalization of several correlation matrices. These techniques are widely used in different applications, like electroencephalogram (EEG) source separation \cite{delorme2012independent}. In the following section brief overviews will be given for each one of the above-mentioned methods. The reader is referred to \cite{naik2011overview, shen2009fast, kong2008review, cardoso1999high, hyvarinen2000independent, langlois2010introduction, choi2005blind, cichocki2002adaptive} and references therein, as well as EEGLAB \cite{delorme2004eeglab} and ICA Central (http://perso.telecom-paristech.fr/$\sim$cardoso/icacentral/)  for further reading regarding other ICA techniques and pointers for implementations. 

\section{Methods}
\label{S:3}
\subsection{InfoMax}

InfoMax is based on a neural network approach \cite{bell1995information}, which tries to maximize the mutual information of the network's output $Y$ about its input $X$, as follows:
\begin{equation}
I(Y,X)=H(Y)-H(Y|X)
\end{equation}
where $H(Y)$ is the entropy of output $Y$ while $H(Y|X)$ represents the entropy of the output that is not a result of input. Considering only the \textit{gradient} of information theoretic quantities with respect to some parameter $w$ in the network, while knowing that $H(Y|X)$ does not depend on $w$, we can have:
\begin{equation}
\frac{\partial}{\partial w}I(X,Y)=\frac{\partial}{\partial w}H(Y)
\end{equation}

Having a neural network with input vector $\mathbf x$, weight matrix $\mathbf W$, bias vector $ \mathbf{w_0}$ and monotonically transformed output vector $\mathbf y= g(\mathbf W \mathbf x) +\mathbf{w_0}$ with $g(u)$ being a sigmoidal function ($g(u)=(1+e^{-u})^{-1}$), the learning rule will be of the form:
\begin{equation}
\bigtriangleup \mathbf W \propto [\mathbf W ^T]^{-1}+(\mathbf 1-2 \mathbf y)\mathbf x^T
\end{equation}
where $\mathbf 1$ is a vector of ones. At each iteration $\mathbf W$ is updated until it reaches convergence. 

\subsection{FastICA}

The differential entropy $H$ of a random vector $y$ with density function $f(y)$ can be defined as:

\begin{equation}
H(y)=-\int f(y) log f(y) dy
\end{equation}
It has been proved that \cite{cover2012elements, papoulis2002probability} a Gaussian variable has the largest entropy among all random variables of equal variance. Therefore to obtain a measure for non-Gaussianity, a modified definition of differential entropy is used which is called \textit{negentropy}:

\begin{equation}
J(y)=H(y_{gauss})-H(y)
\end{equation}
in which $H(y_{gauss})$ is a Gaussian random variable of the same covariance matrix as $y$. This measure is always non-negative since the second term in subtraction is always smaller or equal to the first term. For FastICA, an approximation of the negentropy is used:

\begin{equation}
J(y)\propto [E\{G(y)\}-E\{G(v)\}]^2
\end{equation}
$v$ being a zero-mean and unit variance Gaussian variable and $G$ a non-quadratic function. Two examples for proper function $G$ are $G_1(u)=\frac{1}{a_1} log(cosh(a_1u))$ and $G_2(u)=-exp(-u^2/2)$ for $1\leq a_1\leq 2$. More elaboration on these choices can be found in \cite{hyvarinen2000independent}. Maximization of the non-Gaussianity of $w^T x$ is done using a fixed-point scheme consisting of several steps, with $g$ being the derivative of $G$, as follows:
\begin{enumerate}
\item initialization of the weight vector $w$;
\item $w^+=E\{x(g(w^Tx))\}-E\{g(w^Tx)\}w$;
\item $w=w^+/||w^+||$;
\item \textbf{goto} step 2, until convergence.
\end{enumerate}
In this formulation, the convergence happens when the new and old $w$ vectors are in the same direction which means having their dot product equal to zero. Of course this can happen even if they are in opposite directions too, which is valid since that represents one of he ambiguities of the ICA. 

\subsection{SOBI}

Assuming a more general case of mixed sources with addition of noise $\mathbf n(t)$ we have:
\begin{equation}
\mathbf x(t)=\mathbf A \mathbf s(t)+\mathbf n(t)
\end{equation}
where $\mathbf A$ is a full-ranked complex matrix. The main assumption is to have second order stationary and mutually uncorrelated sources while the additive noise is assumed to be spatially and temporally white and uncorrelated with the source signals. SOBI tries to separate the components by joint diagonalization of several correlation matrices. Based on these assumptions, the algorithm of robust SOBI for $N$ samples consists of several steps as follows \cite{cichocki2002adaptive}:
\begin{enumerate}
\item robust orthogonalization of the sensor signals, $\bar{x}=\mathbf Q x(t)$, $\mathbf Q$ being the orthogonalization matrix;
\item estimate the set of covariance matrices: $$\hat{\mathbf R}_{\bar{x}}(p_i)=\frac{1}{N}\sum_{k=1}^N\bar{x}(t)\bar{x}^T(t-p_i)=\mathbf Q\hat{\mathbf R}_x(p_i) \mathbf Q^T$$ for a set of time lags $(p_1, p_2, ..., p_L)$;
\item performing joint approximate diagonalization: $\mathbf R_{\bar{x}}(p_i)=\mathbf U \mathbf D_i \mathbf U^T, \forall i$, which means estimating the orthogonal matrix $\mathbf U$;
\item estimating the source signals $$\hat{s}(t)= \mathbf U^T \mathbf Q x(t)$$ and the mixing matrix $$\hat{\mathbf W}=\mathbf Q^+ \mathbf U.$$
\end{enumerate}

\subsection{JADE}

The JADE (Joint Approximate Diagonalization of Eigenmatrices) is a natural extension of the SOBI. The algorithm consists of the following steps \cite{cichocki2002adaptive}:
\begin{enumerate}
\item robust pre-whitening or orthogonalization: $\bar{x}=\mathbf Q x(t)$;
\item performing eigen value decomposition (EVD) of the sampled contracted quadri-covariance matrix 

\begin{equation}
\begin{split}
\mathbf C_{\bar{x}}(\mathbf I)=\frac{1}{N}\sum_{k=1}^N\{\bar{x}^T(t)\bar{x}(t)\bar{x}(t)\bar{x}^T(t)\}-2\hat{\mathbf R}_{\bar{x}}(0)\hat{\mathbf R}_{\bar{x}}(0)-tr(\hat{\mathbf R}_{\bar{x}}(0))\hat{\mathbf R}_{\bar{x}}(0)\\=\hat{\mathbf U} \mathbf \Lambda_I\hat{\mathbf U}^T
\end{split}
\end{equation}
with $\hat{\mathbf R}_{\bar{x}}(0)=\frac{1}{N}\sum_{k=1}^N\{\bar{x}(t)\bar{x}^T(t)]\}$ and $\hat{\mathbf U}=[\hat{\mathbf u}_1, \hat{\mathbf u}_2, ..., \hat{\mathbf u}_n]$;
\item estimating the $n$ sampled contracted quadri-covariance matrices:
\begin{equation}
\begin{split}
\mathbf C_{\bar{x}}(\mathbf E_p)=\frac{1}{N}\sum_{k=1}^N\{\bar{x}^T(t) \mathbf E_p\bar{x}(t)\bar{x}(t)\bar{x}^T(t)\}-\hat{\mathbf R}_{\bar{x}}(0)\mathbf E_p\hat{\mathbf R}_{\bar{x}}(0)\\-tr(\mathbf E_p \hat{\mathbf R}_{\bar{x}}(0))\hat{\mathbf R}_{\bar{x}}(0)-\hat{\mathbf R}_{\bar{x}} \mathbf E_p^T\hat{\mathbf R}_{\bar{x}}
\end{split}
\end{equation}
for $\mathbf E_p=\hat{\mathbf u}_p\hat{\mathbf u}_p^T, p=1,2,...,n$;
\item finding an orthogonal joint diagonalization matrix $\mathbf U$ for all $n$ matrices;
\item estimating the mixing matrix using $\hat{\mathbf W}=\mathbf Q^+  \mathbf U$.
\end{enumerate}

For more elaboration on the methods for joint approximate diagonalization, the reader is referred to \cite{cichocki2002adaptive}.

\section{Results and Discussion}
\label{S:4}

For testing the methods, multiple B-scans of the same cross-section in retina are acquired using a Bioptigen OCT device (Bioptigen, Inc. Durham, NC 27709) available at AOIP (Advanced Ocular Imaging Program, Medical College of Wisconsin, Milwaukee, WI). Even though the subject is asked to fixate, the involuntary eye movements (tremor, drifts, micro-saccades) cause misalignment between consecutive B-scans. Considering the effect of eye movements within each B-scan as negligible, a parametric registration technique, namely rigid registration, is suitable for compensating the motion between B-scans. For this, the rigid registration technique implemented in ImageJ \cite{schneider2012nih} is used. After aligning the B-scans, they are vectorized and aggregated into a data matrix and then ICA techniques, implemented in MATLAB (The MathWorks, Inc, Natick, MA 01760), are applied to extract the independent components in the data matrix. The main component is the noise-free image, while the rest are only containing noise. 

For assessing the performance of the methods, several metrics are considered. Considering 9 regions of interest (ROI) as depicted in Fig. 1 in the final result, one only containing background noise and the rest containing image features and homogeneous regions, the metrics can be defined as follows:
\begin{equation}
\begin{split}
SNR_m=20\times log_{10}(\frac{\mu _m}{\sigma _b}) \\
 CNR_m=\frac{\mu _m -\mu _b}{\sqrt{\sigma ^2_m+\sigma ^2_b}} \\
 ENL_m=\frac{\mu_m^2}{\sigma_m^2}
\end{split}
\end{equation}
where $ \mu_b $ and $ \sigma_b $ are the mean and standard deviation of the background noise and $ \mu_m $ and $ \sigma_m $ are the mean and standard deviation of the $ m $-th  ROI containing image features. ENL provides a measure for smoothness of homogeneous regions while SNR and CNR provide information on the signal and contrast of image features with respect to the background noise. The average of these metrics are considered here for comparison. Different numbers of images of the human retina in the central foveal region are considered for assessing the performance of the proposed algorithm. 

Fig. 2 displays the improvement of SNR for different methods using different numbers of input B-scans. Theoretically, having $N$ images, the SNR can be improved by $\sqrt[]{N}$ using average/median filtering. As is obvious, this is not the case for ICA techniques. While all of the ICA techniques outperform median filtering at first, for using up to 20 input B-scans, only the SOBI remains superior for the rest and RUNICA, JADE and FastICA perform poorly. As mentioned before, one of the main constraints on the input data for being able to use ICA is having uncorrelated non-Gaussian components. This can't be satisfied, especially when using excessive number of images. The same pattern can be seen in Fig. 3 and Fig. 4 for the improvement of CNR and ENL, respectively. In case of SOBI, the assumption of having temporally correlated sources in the design of the algorithm makes it better in terms of SNR, CNR and ENL since the input images are acquired from the same location in retina, with the possibility of misalignment due to eye movement.

As for the computational complexity, while JADE shows a quadratic behavior for convergence, the rest show linear increment when having more input B-scans. Fig. 5 shows the graphs for computational time of different methods for different number of input images. Fig. 6 displays close-ups of the original and filtered version of the input images using different techniques for comparison. 

\begin{figure}
\centering
\includegraphics[scale=.55]{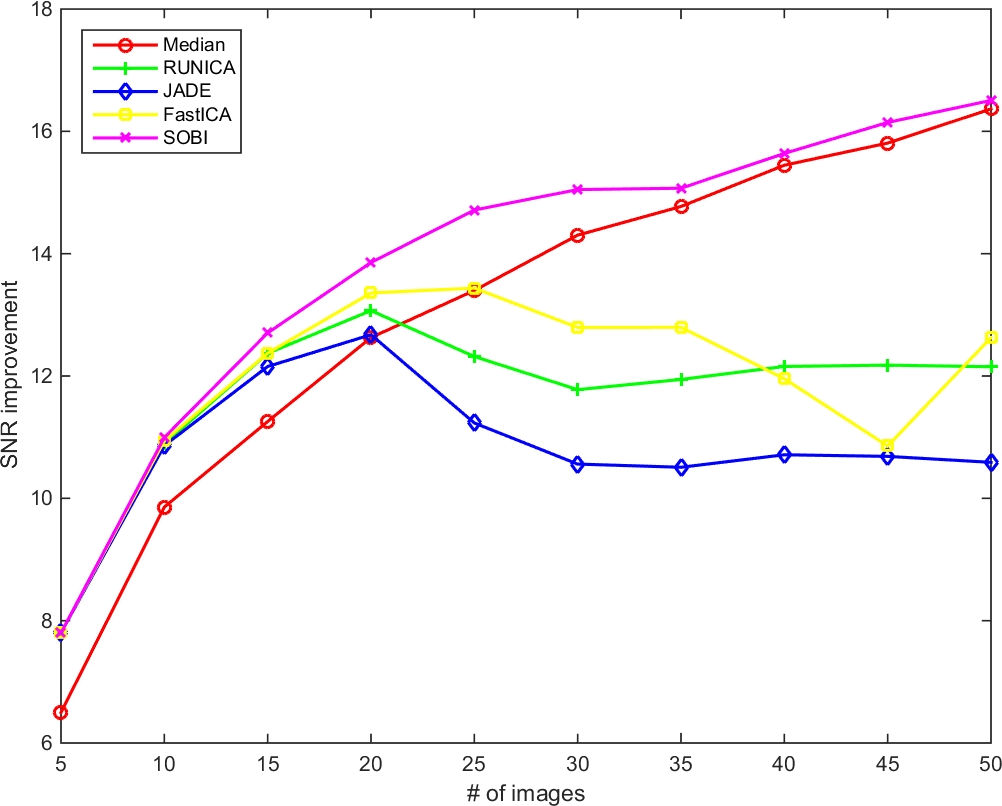}
\caption{SNR improvement for different number of input images (5-50).}
\end{figure}

\begin{figure}
\centering
\includegraphics[scale=.55]{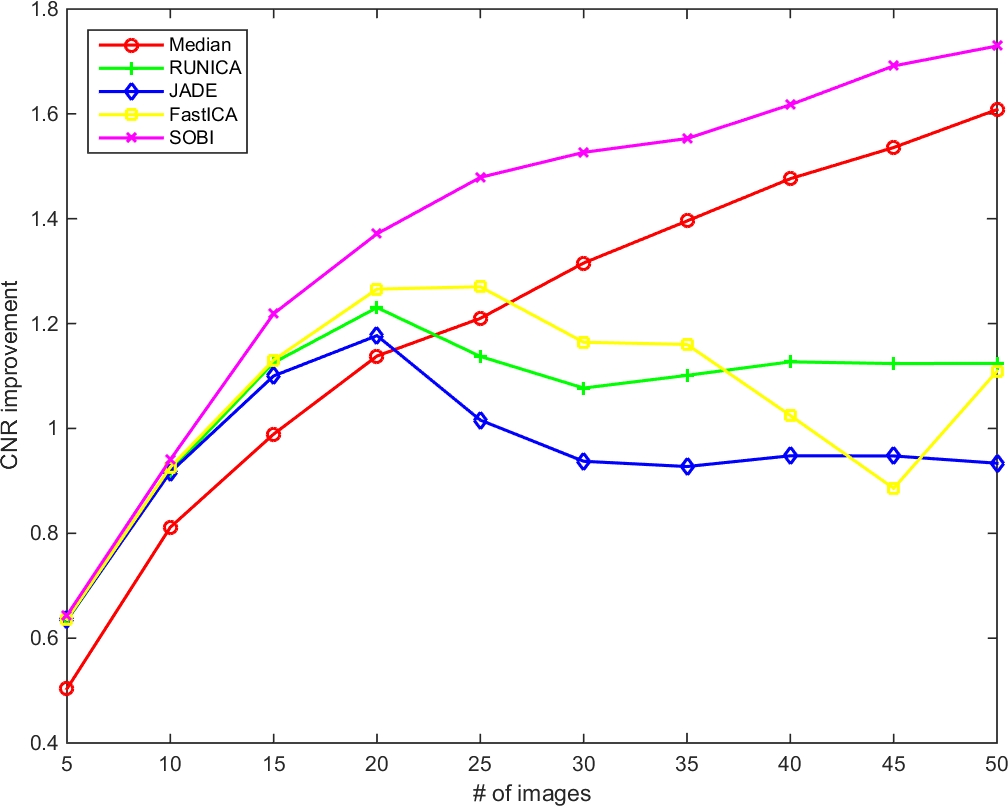}
\caption{CNR improvement for different number of input images (5-50).}
\end{figure}

\begin{figure}
\centering
\includegraphics[scale=.55]{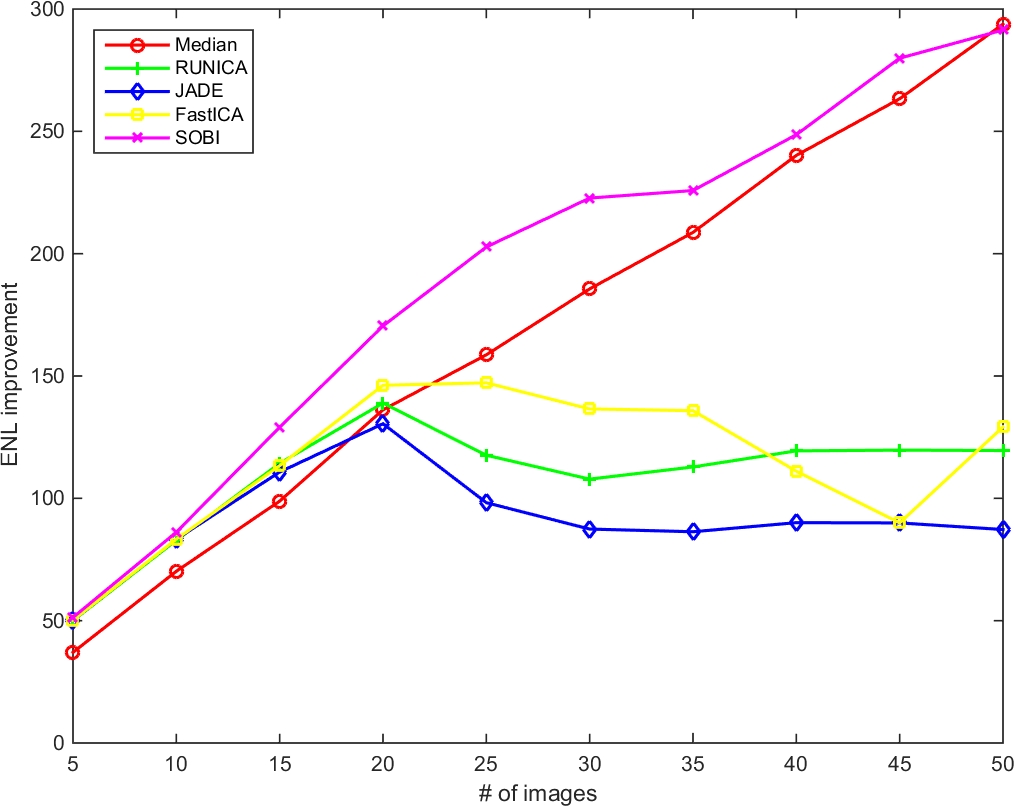}
\caption{ENL improvement for different number of input images (5-50).}
\end{figure}

\begin{figure}
\centering
\includegraphics[scale=.55]{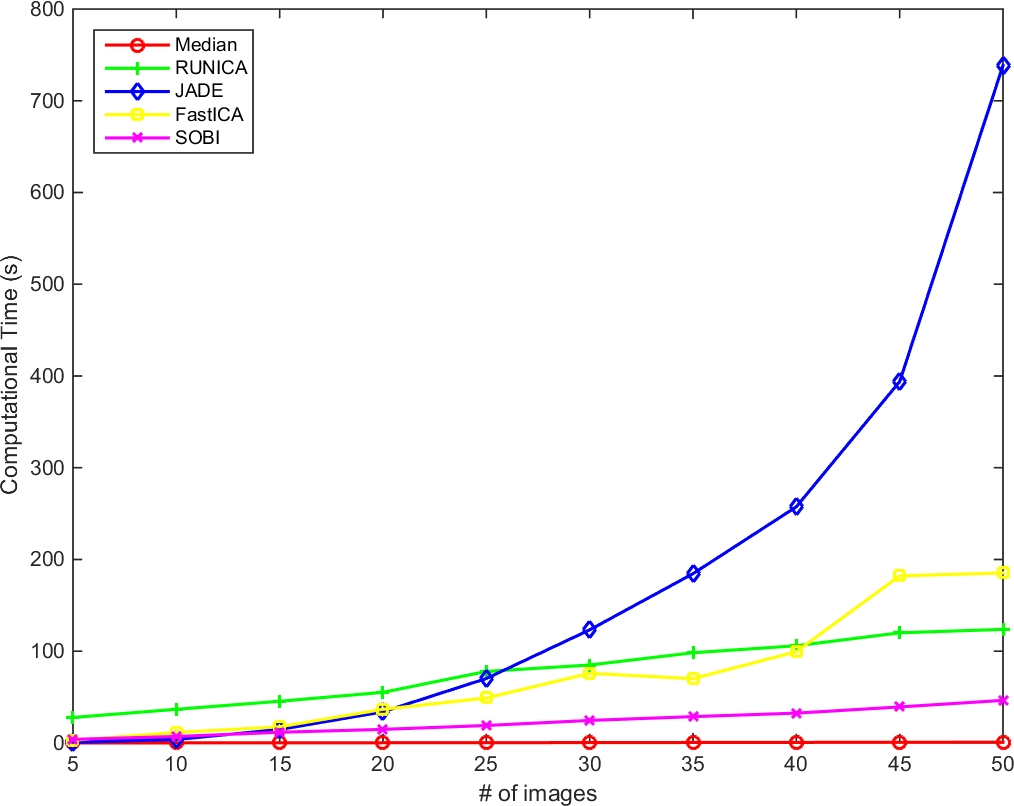}
\caption{Comparison of computational times.}
\end{figure}

\section{Conclusion}
\label{S:5}
In this paper, a new application of ICA techniques for noise reduction of retinal OCT images is proposed. Having a set of B-scans from the same location (to some extent) in retina, the eye movement is compensated using a parametric registration technique for compensating translation and rotation between consecutive B-scans. Then, taking advantage of four ICA techniques, RUNICA, JADE, FastICA and SOBI, the aggregated dataset is analyzed and the noise free image is extracted. Having $N$ B-scans with uncorrelated speckle, the expected improvement in SNR for average/median filtering is $\sqrt[]{N}$. While this is satisfied for up to 20 input B-scans, with ICA techniques outperforming the median filtering, having more input B-scans results in poor performance in all of the ICA methods, except for SOBI. As for the comparison of the computational time, JADE has a quadratic behavior, while the rest show linear increment in computational time when having more images. Overall, SOBI is the best among the ICA techniques considered here in terms of performance based on SNR, CNR and ENL, while needing less computational power. As for pointers towards next possible areas for research, newer techniques for ICA/BSS as well as more analysis on the speckle noise model in OCT images should be explored and investigated, while reducing the computational complexity and exploring the possibility of using the techniques in an on-line manner can be considered as next steps.

\begin{figure}
\centering
\begin{subfigure}[b]{0.125\textwidth}
\centering
\includegraphics[scale=.5]{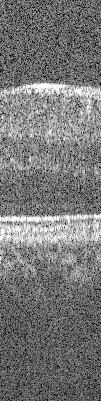}
\caption{}
\end{subfigure}
\begin{subfigure}[b]{0.125\textwidth}
\centering
\includegraphics[scale=.5]{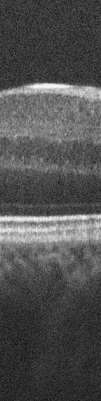}
\caption{}
\end{subfigure}
\begin{subfigure}[b]{0.125\textwidth}
\centering
\includegraphics[scale=.5]{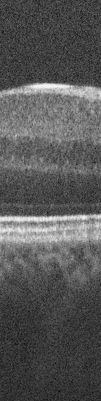}
\caption{}
\end{subfigure}
\begin{subfigure}[b]{0.125\textwidth}
\centering
\includegraphics[scale=.5]{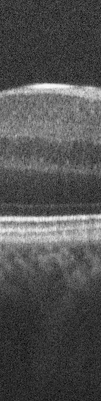}
\caption{}
\end{subfigure}
\begin{subfigure}[b]{0.125\textwidth}
\centering
\includegraphics[scale=.5]{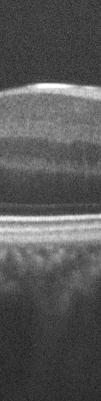}
\caption{}
\end{subfigure}
\caption{A portion of one of the input images (a) and the results of InfoMax (b), JADE (c), FastICA (d) and SOBI (e), resulted from 50 B-scans.}
\end{figure}

\section{Acknowledgment}
The authors would like to thank Dr. Joseph Carroll from Advanced Ocular Imaging Program (AOIP), Medical College of Wisconsin (MCW), Milwaukee, WI for providing the data and insight. This work was partially supported by NIH P30EY001931. Support received by grant 8UL1TR000055 from the Clinical and Translational Science Award (CTSA) program of the National Center for Research Resources and the National Center for Advancing Translational Sciences. Support received by the Clinical and Translational Science
Institute of Southeast Wisconsin through the Advancing a
Healthier Wisconsin endowment of the Medical College of
Wisconsin.




\section{References}
\label{S:6}


\bibliographystyle{model1-num-names}
\bibliography{sample.bib}







\end{document}